%% file: egpaper_final.tex
\documentclass[10pt,twocolumn,letterpaper]{article}

\usepackage{cvpr}
\usepackage{times}
\usepackage{epsfig}
\usepackage{graphicx}
\usepackage{amsmath}
\usepackage{amssymb}
\usepackage{booktabs}
\usepackage{xcolor}
\usepackage{enumitem}
\usepackage{lipsum}

\usepackage[ruled,vlined,linesnumbered]{algorithm2e}
\makeatletter
\newcommand{\algorithmfootnote}[2][\footnotesize]{%
  \let\old@algocf@finish\@algocf@finish
  \def\@algocf@finish{\old@algocf@finish
    \leavevmode\rlap{\begin{minipage}{\linewidth}
    #1#2
    \end{minipage}}%
  }%
}
\makeatother

\definecolor{citecolor}{HTML}{0071bc}
\usepackage[pagebackref=false,breaklinks=true,letterpaper=true,colorlinks,citecolor=citecolor,bookmarks=false]{hyperref}

\newcommand{\app}{\raise.17ex\hbox{$\scriptstyle\sim$}}

\definecolor{codegreen}{rgb}{0.0,0.6,0.0}

\def\cvprPaperID{****} 

\newcommand{\myparagraph}[1]{{\vspace{1.0em} \noindent \bf #1}}

\begin{document}

\title{ByteTrack: Multi-Object Tracking by Associating Every Detection Box}

\author
{
Yifu Zhang$^{1}$, 
~~~
Peize Sun$^{2}$, 
~~~
Yi Jiang$^{3}$, 
~~~
Dongdong Yu$^{3}$, 
~~~
Fucheng Weng$^{1}$, \\
~~~
Zehuan Yuan$^{3}$,
~~~
Ping Luo$^{2}$,
~~~
Wenyu Liu$^{1}$,
~~~
Xinggang Wang$^{1\dag}$
\\[0.2cm]
${^1}$Huazhong University of Science and Technology ~~~
${^2}$The University of Hong Kong ~~~
${^3}$ByteDance Inc. ~~~
}



\maketitle

\let\thefootnote\relax\footnotetext{$^{\dag}$ Corresponding author.}
\let\thefootnote\relax\footnotetext{Part of this work was performed while Yifu Zhang worked as an intern at ByteDance.}

\begin{abstract}
Multi-object tracking (MOT) aims at estimating bounding boxes and identities of objects in videos. Most methods obtain identities by associating detection boxes whose scores are higher than a threshold. The objects with low detection scores, \eg occluded objects, are simply thrown away, which brings non-negligible true object missing and fragmented trajectories. To solve this problem, we present a simple, effective and generic association method, tracking by associating almost every detection box instead of only the high score ones. For the low score detection boxes, we utilize their similarities with tracklets to recover true objects and filter out the background detections. When applied to 9 different state-of-the-art trackers, our method achieves consistent improvement on IDF1 score ranging from 1 to 10 points. To put forwards the state-of-the-art performance of MOT, we design a simple and strong tracker, named ByteTrack. For the first time, we achieve 80.3 MOTA, 77.3 IDF1 and 63.1 HOTA on the test set of MOT17 with 30 FPS running speed on a single V100 GPU. ByteTrack also achieves state-of-the-art performance on MOT20, HiEve and BDD100K tracking benchmarks. The source code, pre-trained models with deploy versions and tutorials of applying to other trackers are released at \url{https://github.com/ifzhang/ByteTrack}.
\end{abstract}

\begin{figure}[!htbp]
    \hspace{-8mm}
	\includegraphics[width=1.2\linewidth]{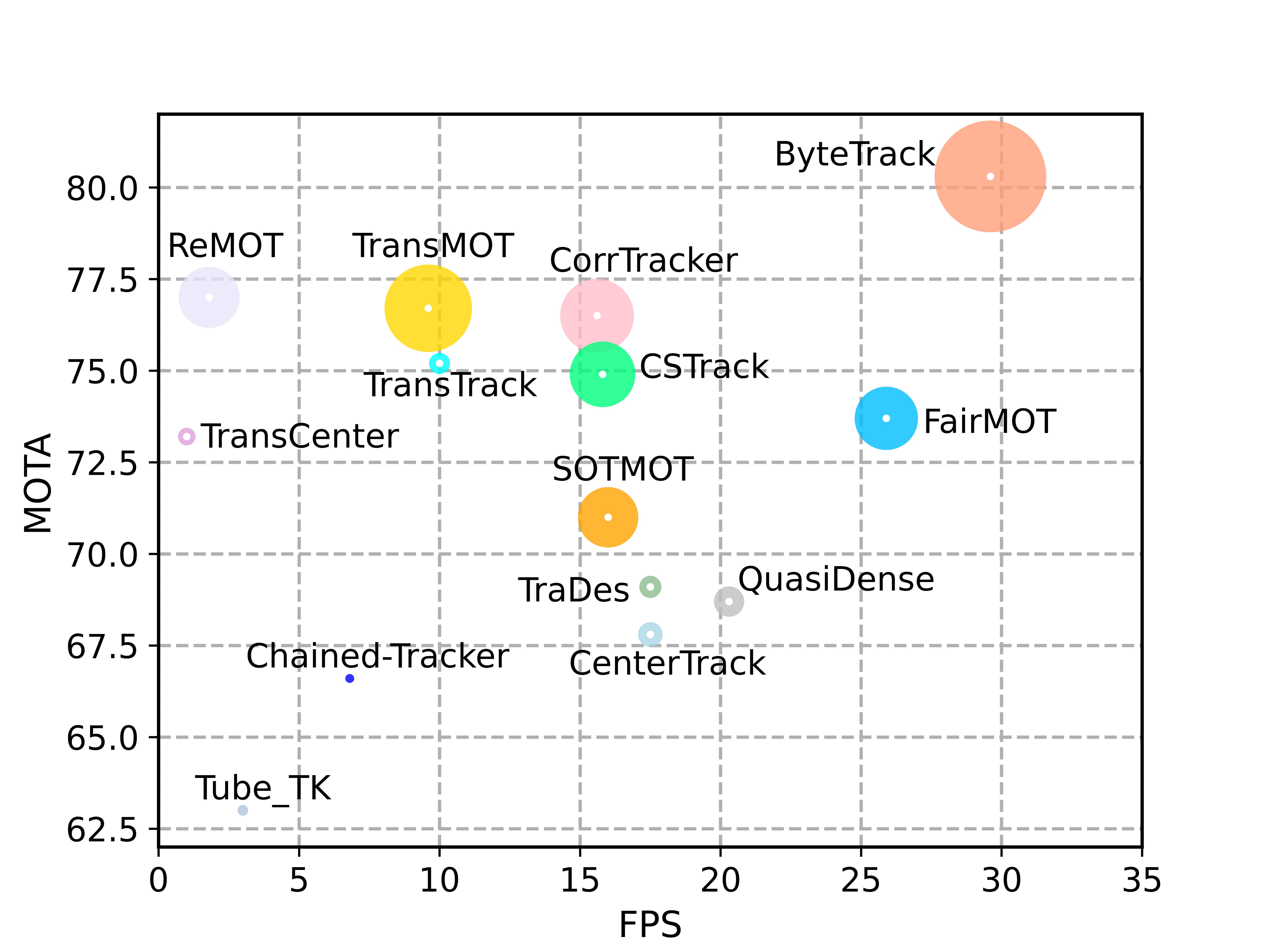}
	\caption{MOTA-IDF1-FPS comparisons of different trackers on the test set of MOT17. The horizontal axis is FPS (running speed), the vertical axis is MOTA, and the radius of circle is IDF1. Our ByteTrack achieves 80.3 MOTA, 77.3 IDF1  on MOT17 test set with 30 FPS running speed, outperforming all previous trackers. Details are given in Table~\ref{table_mot17}.}
	\label{fig:sota}
    \vspace{3mm}
\end{figure}

\begin{figure}[!htbp]
	\centering
	\includegraphics[width=1\linewidth]{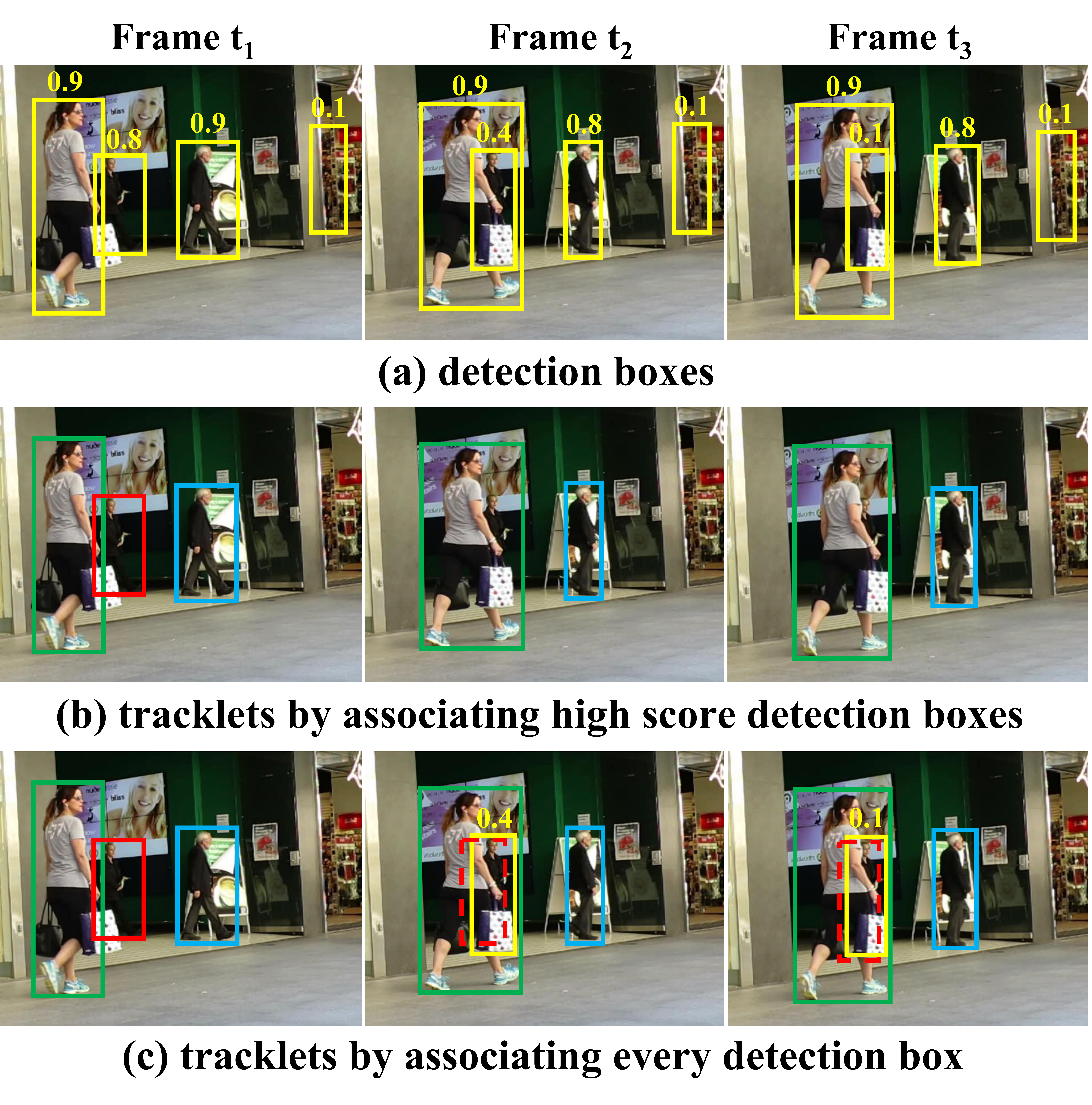}
	\caption{Examples of our method which associates every detection box. (a) shows all the detection boxes with their scores. (b) shows the tracklets obtained by previous methods which associates detection boxes whose scores are higher than a threshold, \ie 0.5. The same box color represents the same identity. (c) shows the tracklets obtained by our method. The dashed boxes represent the predicted box of the previous tracklets using Kalman Filter. The two low score detection boxes are correctly matched to the previous tracklets based on the large IoU. }
	\label{fig:teasing}
    \vspace{3mm}
\end{figure}

\section{Introduction}

\textit{Was vernünftig ist, das ist wirklich; und was wirklich ist, das ist vernünftig.}
\vspace{-2mm}
\begin{flushright}
—— \textit{G. W. F. Hegel}
\end{flushright}

Tracking-by-detection is the most effective paradigm for multi-object tracking (MOT) in current. Due to the complex scenarios in videos, detectors are prone to make imperfect predictions. State-of-the-art MOT methods \cite{berclaz2011multiple,dicle2013way,milan2013continuous,bae2014robust,xiang2015learning,bewley2016simple,wojke2017simple,chen2018real,bergmann2019tracking,zhang2020fairmot,sun2020transtrack} need to deal with true positive / false positive trade-off in detection boxes to eliminate low confidence detection boxes \cite{bernardin2008evaluating,luiten2021hota}. However, is it the right way to eliminate all low confidence detection boxes? Our answer is NO: as Hegel said ``What is reasonable is real; that which is real is reasonable.'' Low confidence detection boxes sometimes indicate the existence of objects, \eg the occluded objects. Filtering out these objects causes irreversible errors for MOT and brings non-negligible missing detection and fragmented trajectories.

Figure~\ref{fig:teasing} (a) and (b) show this problem. In frame $t_1$, we initialize three different tracklets as their scores are all higher than  0.5. However, in frame $t_2$ and frame $t_3$ when occlusion happens, red tracklet's corresponding detection score becomes lower  \ie 0.8 to 0.4 and then 0.4 to 0.1. These detection boxes are eliminated by the thresholding mechanism and the red tracklet disappears accordingly. Nevertheless, if we take every detection box into consideration, more false positives will be introduced immediately, \eg, the most right box in frame $t_3$ of Figure~\ref{fig:teasing} (a). To the best of our knowledge, very few methods \cite{khurana2020detecting,tokmakov2021learning} in MOT are able to handle this detection dilemma.

In this paper, we identify that the similarity with tracklets provides a strong cue to distinguish the objects and background in low score detection boxes. As shown in Figure \ref{fig:teasing} (c), two low score detection boxes are matched to the tracklets by the motion model's predicted boxes, and thus the objects are correctly recovered. At the same time, the background box is removed since it has no matched tracklet.

For making full use of detection boxes from high scores to low ones in the matching process, we present a simple and effective association method BYTE, named for each detection box is a basic unit of the tracklet, as byte in computer program, and our tracking method values every detailed detection box. We first match the high score detection boxes to the tracklets based on motion similarity or appearance similarity. Similar to \cite{bewley2016simple}, we adopt Kalman filter \cite{kalman1960new} to predict the location of the tracklets in the new frame. The similarity can be computed by the IoU or Re-ID feature distance of the predicted box and the detection box. Figure \ref{fig:teasing} (b) is exactly the results after the first matching. Then, we perform the second matching between the unmatched tracklets, \ie the tracklet in red box, and the low score detection boxes using the same motion similarity. Figure \ref{fig:teasing} (c) shows the results after the second matching. The occluded person with low detection scores is matched correctly to the previous tracklet and the background (in the right part of the image) is removed. 

As the integrating topic of object detection and association, a desirable solution to MOT is never a detector and the following association; besides, well-designed of their junction area is also important. The innovation of BYTE lies in the junction area of detection and association, where low score detection boxes are bridges to boost both of them. Benefiting from this integration innovation, when BYTE is applied to 9 different state-of-the-art trackers, including the Re-ID-based ones \cite{wang2020towards,zhang2020fairmot,liang2020rethinking,pang2021quasi}, motion-based ones \cite{zhou2020tracking,wu2021track}, chain-based one \cite{peng2020chained} and attention-based ones \cite{sun2020transtrack,zeng2021motr}, notable improvements are achieved on almost all the metrics including MOTA, IDF1 score and ID switches. For example, we increase the MOTA of CenterTrack \cite{zhou2020tracking} from 66.1 to 67.4, IDF1 from 64.2 to 74.0 and decrease the IDs from 528 to 144 on the half validation set of MOT17.

Towards pushing forwards the state-of-the-art performance of MOT, we propose a simple and strong tracker, named ByteTrack. We adopt a recent high-performance detector YOLOX \cite{ge2021yolox} to obtain the detection boxes and associate them with our proposed BYTE. On the MOT challenges, ByteTrack ranks 1st on both MOT17 \cite{milan2016mot16} and MOT20 \cite{dendorfer2020mot20}, achieving 80.3 MOTA, 77.3 IDF1 and 63.1 HOTA with 30 FPS running speed on V100 GPU on MOT17 and 77.8 MOTA, 75.2 IDF1 and 61.3 HOTA on much more crowded MOT20. ByteTrack also achieves state-of-the-art performance on HiEve \cite{lin2020human} and BDD100K \cite{yu2020bdd100k} tracking benchmarks. We hope the efficiency and simplicity of ByteTrack could make it attractive in real applications such as social computing.

\section{Related Work}
\subsection{Object Detection in MOT}
Object detection is one of the most active topics in computer vision and it is the basis of multi-object tracking. The MOT17 dataset \cite{milan2016mot16} provides detection results obtained by popular detectors such as DPM \cite{felzenszwalb2008discriminatively}, Faster R-CNN \cite{ren2015faster} and SDP \cite{yang2016exploit}. A large number of methods \cite{xu2019spatial,chu2019famnet,bergmann2019tracking,chen2018real,zhu2018online,braso2020learning,hornakova2020lifted} focus on improving the tracking performance based on these given detection results. 

\myparagraph{Tracking by detection.}
With the rapid development of object detection \cite{ren2015faster,he2017mask,redmon2018yolov3,lin2017focal,cai2018cascade,fu2020model,sun2021sparse,peize2020onenet}, more and more methods begin to utilize more powerful detectors to obtain higher tracking performance. The one-stage object detector RetinaNet \cite{lin2017focal} begin to be adopted by several methods such as \cite{lu2020retinatrack,peng2020chained}. CenterNet \cite{zhou2019objects} is the most popular detector adopted by most methods \cite{zhou2020tracking,zhang2020fairmot,wu2021track,zheng2021improving,wang2020joint,tokmakov2021learning,wang2021multiple} for its simplicity and efficiency. The YOLO series detectors \cite{redmon2018yolov3,bochkovskiy2020yolov4} are also adopted by a large number of methods \cite{wang2020towards,liang2020rethinking,liang2021one,chu2021transmot} for its excellent balance of accuracy and speed. Most of these methods directly use the detection boxes on a single image for tracking. 


However, the number of missing detections and very low scoring detections begin to increase when occlusion or motion blur happens in the video sequence, as is pointed out by video object detection methods \cite{tang2019object,luo2019detect}. Therefore, the information of the previous frames are usually leveraged to enhance the video detection performance.

\myparagraph{Detection by tracking.}
Tracking can also adopted to help obtain more accurate detection boxes. Some methods \cite{sanchez2016online,zhu2018online,chu2019famnet,chu2019online,chu2021transmot,chen2018real} utilize single object tracking (SOT) \cite{bertinetto2016fully} or Kalman filter \cite{kalman1960new}  to predict the location of the tracklets in the following frame and fuse the predicted boxes with the detection boxes to enhance the detection results. Other methods \cite{zhang2018integrated,liang2021one} leverage tracked boxes in the previous frames to enhance feature representation of the following frame. Recently, Transformer-based \cite{vaswani2017attention,dosovitskiy2020vit,wang2021pvt,liu2021swin} detectors \cite{carion2020end,zhu2020deformable} are adopted by several methods \cite{sun2020transtrack,meinhardt2021trackformer,zeng2021motr} for its strong ability to propagate boxes between frames. Our method also utilize the similarity with tracklets to strength the reliability of detection boxes. 

After obtaining the detection boxes by various detectors, most MOT methods \cite{wang2020towards,zhang2020fairmot,pang2021quasi,lu2020retinatrack,liang2020rethinking,wu2021track,sun2020transtrack} only keep the high score detection boxes by a threshold, \ie 0.5, and use those boxes as the input of data association. This is because the low score detection boxes contain many backgrounds which harm the tracking performance. However, we observe that many occluded objects can be correctly detected but have low scores. To reduce missing detections and keep the persistence of trajectories, we keep all the detection boxes and associate across every of them. 

\subsection{Data Association}
Data association is the core of multi-object tracking, which first computes the similarity between tracklets and detection boxes and leverage different strategies to match them according to the similarity. 

\myparagraph{Similarity metrics.} 
Location, motion and appearance are useful cues for association. SORT \cite{bewley2016simple} combines location and motion cues in a very simple way. It first adopts Kalman filter \cite{kalman1960new} to predict the location of the tracklets in the new frame and then computes the IoU between the detection boxes and the predicted boxes as the similarity. Some recent methods \cite{zhou2020tracking,sun2020transtrack,wu2021track} design networks to learn object motions and achieve more robust results in cases of large camera motion or low frame rate. Location and motion similarity are accurate in the short-range matching. Appearance similarity are helpful in the long-range matching. An object can be re-identified using appearance similarity after being occluded for a long period of time. Appearance similarity can be measured by the cosine similarity of the Re-ID features. DeepSORT \cite{wojke2017simple} adopts a stand-alone Re-ID model to extract appearance features from the detection boxes. Recently, joint detection and Re-ID models \cite{wang2020towards,zhang2020fairmot,liang2020rethinking,lu2020retinatrack,zhang2021voxeltrack,pang2021quasi} becomes more and more popular because of their simplicity and efficiency.

\myparagraph{Matching strategy.} 
After similarity computation, matching strategy assigns identities to the objects. This can be done by Hungarian Algorithm \cite{kuhn1955hungarian} or greedy assignment \cite{zhou2020tracking}. SORT \cite{bewley2016simple} matches the detection boxes to the tracklets by once matching. DeepSORT \cite{wojke2017simple} proposes a cascaded matching strategy which first matches the detection boxes to the most recent tracklets and then to the lost ones. MOTDT \cite{chen2018real} first utilizes appearance similarity to match and then utilize the IoU similarity to match the unmatched tracklets. QDTrack  \cite{pang2021quasi} turns the appearance similarity into probability by a bi-directional softmax operation and adopts a nearest neighbor search to accomplish matching. Attention mechanism \cite{vaswani2017attention} can directly propagate boxes between frames and perform association implicitly. Recent methods such as \cite{meinhardt2021trackformer,zeng2021motr} propose track queries to find the location of the tracked objects in the following frames. The matching is implicitly performed in the attention interaction process without using Hungarian Algorithm.

All these methods focus on how to design better association methods. However, we argue that the way detection boxes are utilized determines the upper bound of data association and we focus on how to make full use of detection boxes from high scores to low ones in the matching process.

\begin{algorithm}[!h]
\SetAlgoLined
\DontPrintSemicolon
\SetNoFillComment
\footnotesize
\KwIn{A video sequence $\texttt{V}$; object detector $\texttt{Det}$; detection score threshold {$\tau$}}
\KwOut{Tracks $\mathcal{T}$ of the video}

Initialization: $\mathcal{T} \leftarrow \emptyset$\;
\For{frame $f_k$ in $\texttt{V}$}{
	\tcc{Figure 2(a)}
	\tcc{predict detection boxes \& scores}
	$\mathcal{D}_k \leftarrow \texttt{Det}(f_k)$ \;
	$\mathcal{D}_{high} \leftarrow \emptyset$ \;
	\textcolor{codegreen}{$\mathcal{D}_{low} \leftarrow \emptyset$} \;
	\For{$d$ in $\mathcal{D}_k$}{
	\If{$d.score > \tau$}{
	$\mathcal{D}_{high} \leftarrow  \mathcal{D}_{high} \cup \{d\}$ \;
	}
	\Else{
	\textcolor{codegreen}{$\mathcal{D}_{low} \leftarrow  \mathcal{D}_{low} \cup \{d\}$ \;
	}}
	}
	
    \BlankLine	
	\BlankLine
	\tcc{predict new locations of tracks}
	\For{$t$ in $\mathcal{T}$}{
	$t \leftarrow \texttt{KalmanFilter}(t)$ \;
	}
	
    \BlankLine
    \BlankLine
	\tcc{Figure 2(b)}
	\tcc{first association}
	Associate $\mathcal{T}$ and $\mathcal{D}_{high}$ using \texttt{Similarity\#1}\;
	$\mathcal{D}_{remain} \leftarrow \text{remaining object boxes from } \mathcal{D}_{high}$ \;
	$\mathcal{T}_{remain} \leftarrow \text{remaining tracks from } \mathcal{T}$ \;
	
	\BlankLine
	\BlankLine
	\tcc{Figure 2(c)}
    \tcc{second association}
	\textcolor{codegreen}{
	Associate $\mathcal{T}_{remain}$ and $\mathcal{D}_{low}$ using \texttt{similarity\#2}\;}
	\textcolor{codegreen}{
	$\mathcal{T}_{re-remain} \leftarrow \text{remaining tracks from } \mathcal{T}_{remain}$ \;}

    \BlankLine
	\BlankLine
	\tcc{delete unmatched tracks}
	$\mathcal{T} \leftarrow \mathcal{T} \setminus \mathcal{T}_{re-remain}$ \;
	
    \BlankLine
	\BlankLine
	\tcc{initialize new tracks}
    \For{$d$ in $\mathcal{D}_{remain}$}{
	$\mathcal{T} \leftarrow  \mathcal{T} \cup \{d\}$ \;
	}
}
Return: $\mathcal{T}$
\caption{Pseudo-code of BYTE.}
\algorithmfootnote{Track rebirth~\cite{wojke2017simple,zhou2020tracking} is not shown in the algorithm for simplicity. In \textcolor{codegreen}{green} is the key of our method. }
\label{algo:byte}
\end{algorithm}

\section{BYTE}
\label{sec:byte}
We propose a simple, effective and generic data association method, BYTE. Different from previous methods \cite{wang2020towards,zhang2020fairmot,liang2020rethinking,pang2021quasi} which only keep the high score detection boxes, we keep almost every detection box and separate them into high score ones and low score ones. We first associate the high score detection boxes to the tracklets. Some tracklets get unmatched because they do not match to an appropriate high score detection box, which usually happens when occlusion, motion blur or size changing occurs. We then associate the low score detection boxes and these unmatched tracklets to recover the objects in low score detection boxes and filter out background, simultaneously. The pseudo-code of BYTE is shown in Algorithm~\ref{algo:byte}.

The input of BYTE is a video sequence $\texttt{V}$, along with an object detector $\texttt{Det}$. We also set a detection score threshold $\tau$. The output of BYTE is the tracks $\mathcal{T}$ of the video and each track contains the bounding box and identity of the object in each frame.

For each frame in the video, we predict the detection boxes and scores using the detector $\texttt{Det}$. We separate all the detection boxes into two parts $\mathcal{D}_{high}$ and $\mathcal{D}_{low}$ according to the detection score threshold $\tau$. For the detection boxes whose scores are higher than $\tau$, we put them into the high score detection boxes $\mathcal{D}_{high}$. For the detection boxes whose scores are lower than $\tau$, we put them into the low score detection boxes $\mathcal{D}_{low}$ (line 3 to 13 in Algorithm~\ref{algo:byte}). 

After separating the low score detection boxes and the high score detection boxes, we adopt Kalman filter to predict the new locations in the current frame of each track in $\mathcal{T}$ (line 14 to 16 in Algorithm~\ref{algo:byte}).

The first association is performed between the high score detection boxes $\mathcal{D}_{high}$ and all the tracks $\mathcal{T}$ (including the lost tracks $\mathcal{T}_{lost}$). \texttt{Similarity\#1} can be computed by either by the IoU or the Re-ID feature distances between the detection boxes $\mathcal{D}_{high}$ and the predicted box of tracks $\mathcal{T}$. Then, we adopt Hungarian Algorithm \cite{kuhn1955hungarian} to finish the matching based on the similarity. We keep the unmatched detections in $\mathcal{D}_{remain}$ and the unmatched tracks in $\mathcal{T}_{remain}$ (line 17 to 19 in Algorithm~\ref{algo:byte}). 

BYTE is highly flexible and can be compatible to other different association methods. For example, when BYTE is combined with FairMOT~\cite{zhang2020fairmot}, Re-ID feature is added into \texttt{* first association *} in Algorithm~\ref{algo:byte}, others are the same. In the experiments, we apply BYTE to 9 different state-of-the-art trackers and achieve notable improvements on almost all the metrics.

The second association is performed between the low score detection boxes $\mathcal{D}_{low}$ and the remaining tracks $\mathcal{T}_{remain}$ after the first association. We keep the unmatched tracks in $\mathcal{T}_{re-remain}$ and just delete all the unmatched low score detection boxes, since we view them as background. (line 20 to 21 in Algorithm~\ref{algo:byte}). We find it important to use IoU alone as the \texttt{Similarity\#2} in the second association because the low score detection boxes usually contains severe occlusion or motion blur and appearance features are not reliable. Thus, when apply BYTE to other Re-ID based trackers \cite{wang2020towards,zhang2020fairmot,pang2021quasi}, we do not adopt appearance similarity in the second association. 

After the association, the unmatched tracks will be deleted from the tracklets. We do not list the procedure of track rebirth \cite{wojke2017simple,chen2018real,zhou2020tracking} in Algorithm~\ref{algo:byte} for simplicity. Actually, it is necessary for the long-range association to preserve the identity of the tracks. For the unmatched tracks $\mathcal{T}_{re-remain}$ after the second association, we put them into $\mathcal{T}_{lost}$. For each track in $\mathcal{T}_{lost}$, only when it exists for more than a certain number of frames, \ie 30, we delete it from the tracks $\mathcal{T}$. Otherwise, we remain the lost tracks $\mathcal{T}_{lost}$ in $\mathcal{T}$(line 22 in Algorithm~\ref{algo:byte}).Finally, we initialize new tracks from the unmatched high score detection boxes $\mathcal{D}_{remain}$ after the first association. (line 23 to 27 in Algorithm~\ref{algo:byte}).The output of each individual frame is the bounding boxes and identities of the tracks $\mathcal{T}$ in the current frame. Note that we do not output the boxes and identities of $\mathcal{T}_{lost}$.

To put forwards the state-of-the-art performance of MOT, we design a simple and strong tracker, named ByteTrack, by equipping the high-performance detector YOLOX \cite{ge2021yolox} with our association method BYTE.

\section{Experiments}

\subsection{Setting}
\myparagraph{Datasets.} 
We evaluate BYTE and ByteTrack on MOT17 \cite{milan2016mot16} and MOT20 \cite{dendorfer2020mot20} datasets under the ``private detection'' protocol. Both datasets contain training sets and test sets, without validation sets. For ablation studies, we use the first half of each video in the training set of MOT17 for training and the last half for validation following \cite{zhou2020tracking}. We train on the combination of CrowdHuman dataset \cite{shao2018crowdhuman} and MOT17 half training set following \cite{zhou2020tracking,sun2020transtrack,zeng2021motr,wu2021track}. We add Cityperson \cite{zhang2017citypersons} and ETHZ \cite{ess2008mobile} for training following \cite{wang2020towards,zhang2020fairmot,liang2020rethinking} when testing on the test set of MOT17. We also test ByteTrack on HiEve \cite{lin2020human} and BDD100K \cite{yu2020bdd100k} datasets. HiEve is a large scale human-centric dataset focusing on crowded and complex events. BDD100K is the largest driving video dataset and the dataset splits of the MOT task are 1400 videos for training, 200 videos for validation and 400 videos for testing. It needs to track objects of 8 classes and contains cases of large camera motion. 

\myparagraph{Metrics.}
We use the CLEAR metrics \cite{bernardin2008evaluating}, including MOTA, FP, FN, IDs, \textit{etc.}, IDF1 \cite{ristani2016performance} and HOTA \cite{luiten2021hota} to evaluate different aspects of the tracking performance. MOTA is computed based on FP, FN and IDs. Considering the amount of FP and FN are larger than IDs, MOTA focuses more on the detection performance. IDF1 evaluates the identity preservation ability and focus more on the association performance. HOTA is a very recently proposed metric which explicitly balances the effect of performing accurate detection, association and localization. For BDD100K dataset, there are some multi-class metrics such as mMOTA and mIDF1. mMOTA / mIDF1 is computed by averaging the MOTA / IDF1 of all the classes. 

\myparagraph{Implementation details.}
For BYTE, the default detection score threshold $\tau$ is 0.6, unless otherwise specified. For the benchmark evaluation of MOT17, MOT20 and HiEve, we only use IoU as the similarity metrics. In the linear assignment step, if the IoU between the detection box and the tracklet box is smaller than 0.2, the matching will be rejected. For the lost tracklets, we keep it for 30 frames in case it appears again. For BDD100K, we use UniTrack \cite{wang2021different} as the Re-ID model. In ablation study, we use FastReID \cite{he2020fastreid} to extract Re-ID features for MOT17. 

For ByteTrack, the detector is YOLOX \cite{ge2021yolox} with YOLOX-X as the backbone and COCO-pretrained model \cite{lin2014microsoft} as the initialized weights. For MOT17, the training schedule is 80 epochs on the combination of MOT17, CrowdHuman, Cityperson and ETHZ. For MOT20 and HiEve, we only add CrowdHuman as additional training data. For BDD100K, we do not use additional training data and only train 50 epochs. The input image size is 1440 $\times $800 and the shortest side ranges from 576 to 1024 during multi-scale training. The data augmentation includes Mosaic \cite{bochkovskiy2020yolov4} and Mixup \cite{zhang2017mixup}. The model is trained on 8 NVIDIA Tesla V100 GPU with batch size of 48. The optimizer is SGD with weight decay of $5\times10^{-4}$ and momentum of 0.9. The initial learning rate is $10^{-3}$ with 1 epoch warm-up and cosine annealing schedule. The total training time is about 12 hours. Following \cite{ge2021yolox}, FPS is measured with FP16-precision \cite{micikevicius2017mixed} and batch size of 1 on a single GPU.

\subsection{Ablation Studies on BYTE}

\myparagraph{Similarity analysis. }
We choose different types of similarity for the first association and the second association of BYTE. The results are shown in Table~\ref{table_sim}. We can see that either IoU or Re-ID can be a good choice for \texttt{Similarity\#1} on MOT17. IoU achieves better MOTA and IDs while Re-ID achieves higher IDF1. On BDD100K, Re-ID achieves much better results than IoU in the first association. This is because BDD100K contains large camera motion and the annotations are in low frame rate, which causes failure of motion cues. It is important to utilize IoU as \texttt{Similarity\#2} in the second association on both datasets because the low score detection boxes usually contains severe occlusion or motion blur and thus Re-ID features are not reliable. From Table~\ref{table_sim} we can find that using IoU as \texttt{Similarity\#2} increases about 1.0 MOTA compared to Re-ID, which indicates that Re-ID features of the low score detection boxes are not reliable. 

\begin{table*}[!h]
\begin{center}
{\input{tables/sim}}
\end{center}
\vspace{-5mm}
\caption{Comparison of different type of similarity metrics used in the first association and the second association of BYTE on MOT17 and BDD100K validation set. The best results are shown in \textbf{bold}.}
\label{table_sim}
\end{table*}

\myparagraph{Comparisons with other association methods.}
We compare BYTE with other popular association methods including SORT \cite{bewley2016simple}, DeepSORT \cite{wojke2017simple} and MOTDT \cite{chen2018real} on the validation set of MOT17 and BDD100K. The results are shown in Table~\ref{table_ass}. 

SORT can be seen as our baseline method because both methods only adopt Kalman filter to predict the object motion.  We can find that BYTE improves the MOTA metric of SORT from 74.6 to 76.6, IDF1 from 76.9 to 79.3 and decreases IDs from 291 to 159. This highlights the importance of the low score detection boxes and proves the ability of BYTE to recover object boxes from low score one.

DeepSORT utilizes additional Re-ID models to enhance the long-range association. We surprisingly find BYTE also has additional gains compared with DeepSORT. This suggests a simple Kalman filter can perform long-range association and achieve better IDF1 and IDs when the detection boxes are accurate enough. We note that in severe occlusion cases, Re-ID features are vulnerable and may lead to identity switches, instead, motion model behaves more reliably.

\begin{table*}[t]
\begin{center}
{\input{tables/ass}}
\end{center}
\vspace{-5mm}
\caption{Comparison of different data association methods on MOT17 and BDD100K validation set. The best results are shown in \textbf{bold}. }
\label{table_ass}
\end{table*}

\begin{figure}[t]
	\centering
	\includegraphics[width=0.95\linewidth]{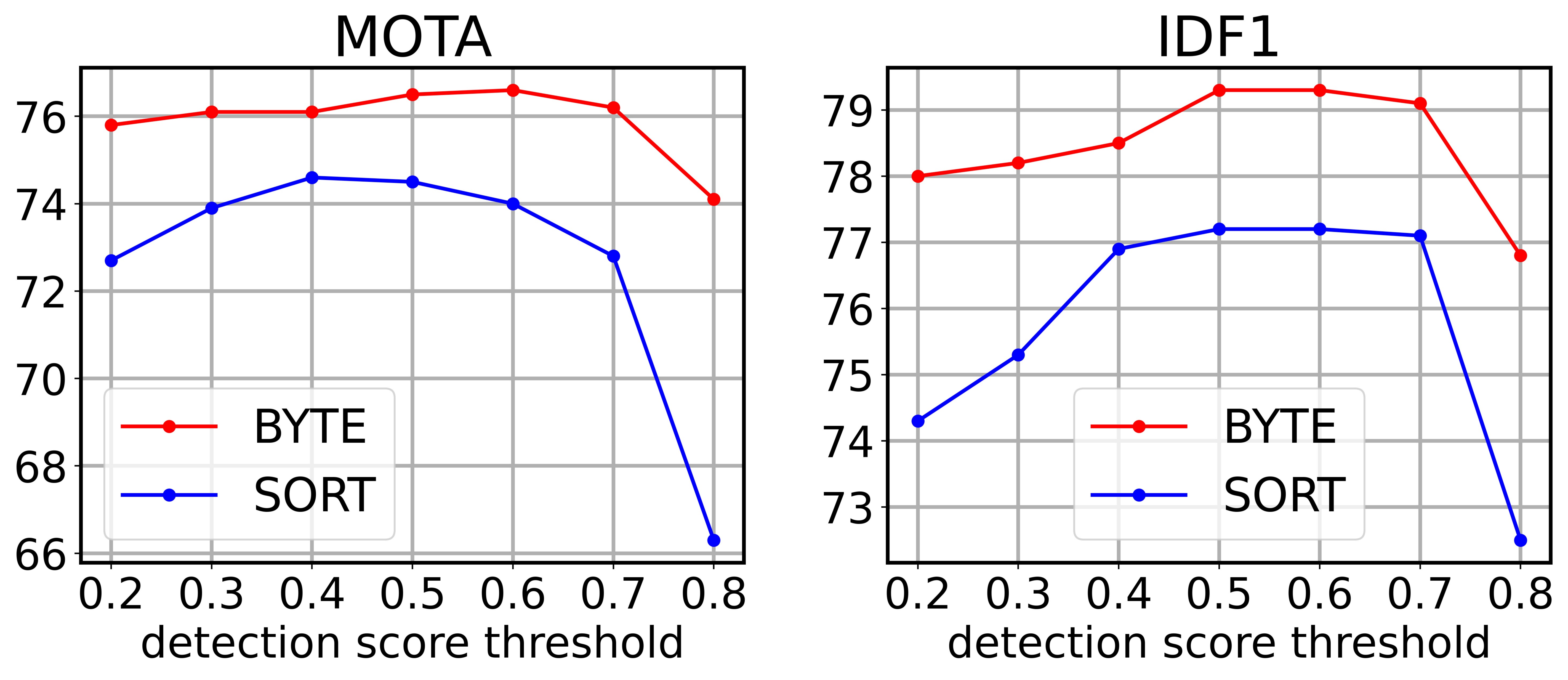}
	\caption{Comparison of the performances of BYTE and SORT under different detection score thresholds. The results are from the validation set of MOT17. }
	\label{fig:threshold}
\end{figure}

MOTDT integrates motion-guided box propagation results along with detection results to associate unreliable detection results with tracklets. Although sharing the similar motivation, MOTDT is behind BYTE by a large margin. We explain that MOTDT uses propagated boxes as tracklet boxes, which may lead to locating drifts in tracking. Instead, BYTE uses low-score detection boxes to re-associate those unmatched tracklets, therefore, tracklet boxes are more accurate.

Table~\ref{table_ass} also shows the results on BDD100K dataset. BYTE also outperforms other association methods by a large margin. Kalman filter fails in autonomous driving scenes and it is the main reason for the low performance of SORT, DeepSORT and MOTDT. Thus, we do not use Kalman filter on BDD100K. Additional off-the-shelf Re-ID models greatly improve the performance of BYTE on BDD100K.

\myparagraph{Robustness to detection score threshold.}
The detection score threshold $\tau_{high}$ is a sensitive hyper-parameter and needs to be carefully tuned in the task of multi-object tracking. We change it from 0.2 to 0.8 and compare the MOTA and IDF1 score of BYTE and SORT. The results are shown in Figure~\ref{fig:threshold}. From the results we can see that BYTE is more robust to the detection score threshold than SORT. This is because the second association in BYTE recovers the objects whose scores are lower than $\tau_{high}$, and thus considers almost every detection box regardless of the change of $\tau_{high}$.

\begin{figure}[t]
    \hspace{-4mm}
	\includegraphics[width=1.05\linewidth]{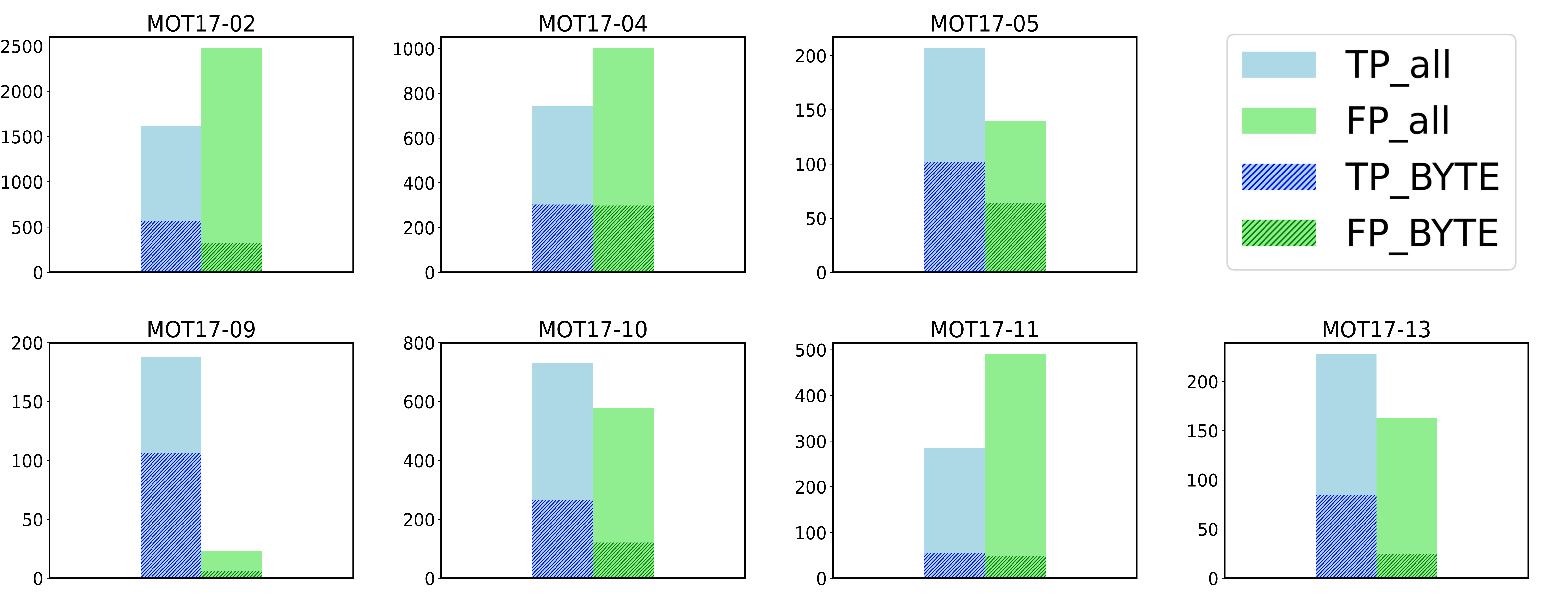}
	\caption{Comparison of the number of TPs and FPs in all low score detection boxes and the low score tracked boxes obtained by BYTE. The results are from the validation set of MOT17. }
	\label{fig:bar}
	\vspace{-1mm}
\end{figure}

\myparagraph{Analysis on low score detection boxes.}
To prove the effectiveness of BYTE, we collect the number of TPs and FPs in the low score boxes obtained by BYTE. We use the half training set of MOT17 and CrowdHuman for training and evaluate on the half validation set of MOT17. First, we keep all the low score detection boxes whose scores range from $\tau_{low}$ to $\tau_{high}$ and classify the TPs and FPs using ground truth annotations. Then, we select the tracking results obtained by BYTE from low score detection boxes. The results of each sequence are shown in Figure~\ref{fig:bar}. We can see that BYTE obtains notably more TPs than FPs from the low score detection boxes even though some sequences (\ie MOT17-02) have much more FPs in all the detection boxes. The obtained TPs notably increases MOTA from 74.6 to 76.6 as is shown in Table~\ref{table_ass}.

\myparagraph{Applications on other trackers.}
We apply BYTE on 9 different state-of-the-arts trackers, including JDE \cite{wang2020towards}, CSTrack \cite{liang2020rethinking}, FairMOT \cite{zhang2020fairmot}, TraDes \cite{wu2021track}, QDTrack  \cite{pang2021quasi}, CenterTrack \cite{zhou2020tracking}, Chained-Tracker \cite{peng2020chained}, TransTrack \cite{sun2020transtrack} and MOTR \cite{zeng2021motr}. Among these trackers, JDE, CSTrack, FairMOT, TraDes adopt a combination of motion and Re-ID similarity. QDTrack  adopts Re-ID similarity alone. CenterTrack and TraDes predict the motion similarity by the learned networks. Chained-Tracker adopts the chain structure and outputs the results of two consecutive frames simultaneously and associate in the same frame by IoU. TransTrack and MOTR adopt the attention mechanism to propagate boxes among frames. Their results are shown in the first line of each tracker in Table~\ref{table_app}.To evaluate the effectiveness of BYTE, we design two different modes to apply BYTE to these trackers.

\begin{itemize}[leftmargin=*]
    \item The first mode is to insert BYTE into the original association methods of different trackers, as is shown in the second line of the results of each tracker in Table~\ref{table_app}. Take FairMOT\cite{zhang2020fairmot} for example, after the original association is done, we select all the unmatched tracklets and associate them with the low score detection boxes following the \texttt{* second association *} in Algorithm~\ref{algo:byte}. Note that for the low score objects, the Re-ID features are not reliable so we only adopt the IoU between the detection boxes and the tracklet boxes after motion prediction as the similarity. We do not apply the first mode of BYTE to Chained-Tracker because we find it is difficult to implement in the chain structure. 
    
    \item The second mode is to directly use the detection boxes of these trackers and associate using the whole procedure in Algorithm~\ref{algo:byte}, as is shown in the third line of the results of each tracker in Table~\ref{table_app}. 
    
\end{itemize}

\begin{table}[t]
\begin{center}
\scalebox{0.75}{\input{tables/app}}
\end{center}
\vspace{-5mm}
\caption{Results of applying BYTE to 9 different state-of-the-art trackers on the MOT17 validation set. ``K'' is short for Kalman Filter. In green are the improvements of at least \textcolor{codegreen}{+\textbf{1.0}} point. }
\label{table_app}
\vspace{-2mm}
\end{table}

We can see that in both modes, BYTE can bring stable improvements over almost all the metrics including MOTA, IDF1 and IDs. For example, BYTE increases CenterTrack by 1.3 MOTA and 9.8 IDF1, Chained-Tracker by 1.9 MOTA and 5.8 IDF1, TransTrack by 1.2 MOTA and 4.1 IDF1. The results in Table~\ref{table_app} indicate that BYTE has strong generalization ability and can be easily applied to existing trackers to obtain performance gain.

\begin{table}[t]
\begin{center}
\scalebox{0.8}{\input{tables/mot17}}
\end{center}
\vspace{-5mm}
\caption{Comparison of the state-of-the-art methods under the “private detector” protocol on MOT17 test set. The best results are shown in \textbf{bold}. MOT17 contains rich scenes and half of the sequences are captured with camera motion. \textbf{ByteTrack} ranks 1st among all the trackers on the leaderboard of MOT17 and outperforms the second one ReMOT by a large margin on almost all the metrics. It also has the highest running speed among all trackers.}
\label{table_mot17}
\end{table}

\begin{table}[t]
\begin{center}
\scalebox{0.8}{\input{tables/mot20}}
\end{center}
\vspace{-5mm}
\caption{Comparison of the state-of-the-art methods under the “private detector” protocol on MOT20 test set. The best results are shown in \textbf{bold}. The scenes in MOT20 are much more crowded than those in MOT17. \textbf{ByteTrack} ranks 1st among all the trackers on the leaderboard of MOT20 and outperforms the second one SOTMOT by a large margin on all the metrics. It also has the highest running speed among all trackers. }
\label{table_mot20}
\end{table}

\begin{table}[t]
\begin{center}
\scalebox{0.8}{\input{tables/hie}}
\end{center}
\vspace{-5mm}
\caption{Comparison of the state-of-the-art methods under the “private detector” protocol on HiEve test set. The best results are shown in \textbf{bold}. HiEve has more complex events than MOT17 and MOT20. \textbf{ByteTrack} ranks 1st among all the trackers on the leaderboard of HiEve and outperforms the second one CenterTrack by a large margin on all the metrics. }
\label{table_hie}
\end{table}

\begin{table*}[t]
\begin{center}
\scalebox{0.9}{\input{tables/bdd100k}}
\end{center}
\vspace{-5mm}
\caption{Comparison of the state-of-the-art methods on BDD100K test set. The best results are shown in \textbf{bold}. \textbf{ByteTrack} ranks 1st among all the trackers on the leaderboard of BDD100K and outperforms the second one QDTrack by a large margin on most metrics. }
\label{table_bdd}
\vspace{-4mm}
\end{table*}

\subsection{Benchmark Evaluation}
We compare ByteTrack with the state-of-the-art trackers on the test set of MOT17, MOT20 and HiEve under the private detection protocol in Table~\ref{table_mot17}, Table~\ref{table_mot20} and Table~\ref{table_hie}, respectively. All the results are directly obtained from the official MOT Challenge evaluation server\footnote{\url{https://motchallenge.net}} and the Human in Events server\footnote{\url{http://humaninevents.org}}.

\myparagraph{MOT17.}
ByteTrack ranks 1st among all the trackers on the leaderboard of MOT17. Not only does it achieve the best accuracy (\ie 80.3 MOTA, 77.3 IDF1 and 63.1 HOTA), but also runs with highest running speed (30 FPS). It outperforms the second-performance tracker \cite{yang2021remot} by a large margin (\ie +3.3 MOTA, +5.3 IDF1 and +3.4 HOTA). Also, we use less training data than many high performance methods such as \cite{zhang2020fairmot,liang2020rethinking,wang2021multiple,shan2020tracklets,liang2021one} (29K images vs. 73K images). It is worth noting that we only leverage the simplest similarity computation method Kalman filter in the association step compared to other methods \cite{zhang2020fairmot,liang2020rethinking,pang2021quasi,wang2020joint,zeng2021motr,sun2020transtrack} which additionally adopt Re-ID similarity or attention mechanisms. All these indicate that ByteTrack is a simple and strong tracker.

\myparagraph{MOT20.}
Compared with MOT17, MOT20 has much more crowded scenarios and occlusion cases. The average number of pedestrians in an image is 170 in the test set of MOT20. ByteTrack also ranks 1st among all the trackers on the leaderboard of MOT20 and outperforms other trackers by a large margin on almost all the metrics. For example, it increases MOTA from 68.6 to 77.8, IDF1 from 71.4 to 75.2 and decreases IDs by 71\% from 4209 to 1223. It is worth noting that ByteTrack achieves extremely low identity switches, which further indicates that associating every detection boxes is very effective under occlusion cases. 

\myparagraph{Human in Events.}
Compared with MOT17 and MOT20, HiEve contains more complex events and more diverse camera views. We train ByteTrack on CrowdHuman dataset and the training set of HiEve. ByteTrack also ranks 1st among all the trackers on the leaderboard of HiEve and outperforms other state-of-the-art trackers by a large margin. For example, it increases MOTA from 40.9 to 61.3 and IDF1 from 45.1 to 62.9. The superior results indicate that ByteTrack is robust to complex scenes. 

\myparagraph{BDD100K.}
BDD100K is multiple categories tracking dataset in autonomous driving scenes. The challenges include low frame rate and large camera motion. We utilize a simple ResNet-50 ImageNet classification model from UniTrack \cite{wang2021different} to extract Re-ID features and compute appearance similarity. ByteTrack ranks first on the leaderboard of BDD100K. It increases mMOTA from 36.6 to 45.5 on the validation set and 35.5 to 40.1 on the test set, which indicates that ByteTrack can also handle the challenges in autonomous driving scenes.

\section{Conclusion}
We present a simple yet effective data association method BYTE for multi-object tracking. BYTE can be easily applied to existing trackers and achieve consistent improvements. We also propose a strong tracker ByteTrack, which achieves 80.3 MOTA, 77.3 IDF1 and 63.1 HOTA on MOT17 test set with 30 FPS, ranking 1st among all the trackers on the leaderboard.
ByteTrack is very robust to occlusion for its accurate detection performance and the help of associating low score detection boxes. It also sheds light on how to make the best use of detection results to enhance multi-object tracking. We hope the high accuracy, fast speed and simplicity of ByteTrack can make it attractive in real applications.


\appendix


\section{Bounding box annotations}

We note MOT17 \cite{milan2016mot16} requires the bounding boxes \cite{zhou2020tracking} covering the whole body, even though the object is occluded or partly out of the image. However, the default implementation of YOLOX clips the detection boxes inside the image area. To avoid the wrong detection results around the image boundary, we modify YOLOX in terms of data pre-processing and label assignment. We do not clip the bounding boxes inside the image during the data pre-processing and data augmentation procedure. We only delete the boxes which are fully outside the image after data augmentation. In the SimOTA label assignment strategy, the positive samples need to be around the center of the object, while the center of the whole body boxes may lie out of the image, so we clip the center of the object inside the image. 

MOT20 \cite{dendorfer2020mot20}, HiEve \cite{lin2020human} and BDD100K clip the bounding box annotations inside the image in and thus we just use the original setting of YOLOX.

\section{Tracking performance of light models}
We compare BYTE and DeepSORT \cite{wojke2017simple} using light detection models. We use YOLOX \cite{ge2021yolox} with different backbones as our detector. All models are trained on CrowdHuman and the half training set of MOT17. The input image size is $1088 \times 608$ and the shortest side ranges from 384 to 832 during multi-scale training. The results are shown in Table~\ref{table_light}. We can see that BYTE brings stable improvements on MOTA and IDF1 compared to DeepSORT, which indicates that BYTE is robust to detection performance. It is worth noting that when using YOLOX-Nano as backbone, BYTE brings 3 points higher MOTA than DeepSORT, which makes it more appealing in real applications. 

\begin{table}[h]
\begin{center}
\scalebox{0.75}{\input{tables/light}}
\end{center}
\vspace{-5.5mm}
\caption{Comparison of BYTE and DeepSORT using light detection models on the MOT17 validation set. }
\label{table_light}
\vspace{-4mm}
\end{table}

\begin{table}[h]
\begin{center}
{\input{tables/input_size}}
\end{center}
\vspace{-5.5mm}
\caption{Comparison of different input sizes on the MOT17 validation set. The total running time is a combination of the detection time and the association time. The best results are shown in \textbf{bold}. }
\label{table_size}
\vspace{-4mm}
\end{table}

\begin{table}[h]
\begin{center}
{\input{tables/training_data}}
\end{center}
\vspace{-5.5mm}
\caption{Comparison of different training data on the MOT17 validation set. ``MOT17'' is short for the MOT17 half training set. ``CH'' is short for the CrowdHuman dataset. ``CE'' is short for the Cityperson and ETHZ datasets. The best results are shown in \textbf{bold}. }
\label{table_data}
\vspace{-4mm}
\end{table}

\section{Ablation Studies on ByteTrack}
\noindent \textbf{Speed v.s. accuracy.}
We evaluate the speed and accuracy of ByteTrack using different size of input images during inference. All experiments use the same multi-scale training. The results are shown in Table~\ref{table_size}. The input size during inference ranges from $512 \times 928$ to $800 \times 1440$. The running time of the detector ranges from 17.9 ms to 30.0 ms and the association time is all around 4.0 ms. ByteTrack can achieve 75.0 MOTA with 45.7 FPS running speed and 76.6 MOTA with 29.6 FPS running speed, which has advantages in practical applications. 

\noindent \textbf{Training data.}
We evaluate ByteTrack on the half validation set of MOT17 using different combinations of training data. The results are shown in Table~\ref{table_data}. When only using the half training set of MOT17, the performance achieves 75.8 MOTA, which already outperforms most methods. This is because we use strong augmentations such as Mosaic \cite{bochkovskiy2020yolov4} and Mixup \cite{zhang2017mixup}. When further adding CrowdHuman, Cityperson and ETHZ for training, we can achieve 76.7 MOTA and 79.7 IDF1. The big improvement of IDF1 arises from that the CrowdHuman dataset can boost the detector to recognize occluded person, therefore, making the Kalman Filter generate smoother predictions and enhance the association ability of the tracker. 

The experiments on training data suggest that ByteTrack is not data hungry. This is a big advantage for real applications, comparing with previous methods \cite{zhang2020fairmot,liang2020rethinking,wang2021multiple,liang2021one} that require more than 7 data sources \cite{milan2016mot16,ess2008mobile,zhang2017citypersons,xiao2017joint,zheng2017person,dollar2009pedestrian,shao2018crowdhuman} to achieve high performance.

\begin{table}[t]
\begin{center}
{\input{tables/inter}}
\end{center}
\vspace{-5.5mm}
\caption{Comparison of different interpolation intervals on the MOT17 validation set. The best results are shown in \textbf{bold}.}
\label{table_inter}
\vspace{-4mm}
\end{table}

\section{Tracklet interpolation}

We notice that there are some fully-occluded pedestrians in MOT17, whose visible ratio is 0 in the ground truth annotations. Since it is almost impossible to detect them by visual cues, we obtain these objects by tracklet interpolation. 

Suppose we have a tracklet $T$, its tracklet box is lost due to occlusion from frame $t_1$ to $t_2$. The tracklet box of $T$ at frame $t_1$ is $B_{t_1} \in \mathbb{R}^4$ which contains the top left and bottom right coordinate of the bounding box. Let $B_{t_2}$ represent the tracklet box of $T$ at frame $t_2$. We set a hyper-parameter $\sigma$ representing the max interval we perform tracklet interpolation, which means tracklet interpolation is performed when $t_2 - t_1 \leq \sigma$, . The interpolated box of tracklet $T$ at frame t can be computed as follows:
\begin{equation}
    B_t = B_{t_1} + (B_{t_2} - B_{t_1}) \frac{t - t_1}{t_2 - t_1},
\end{equation}
where $t_1 < t < t_2$. 

As shown in Table~\ref{table_inter}, tracklet interpolation can improve MOTA from 76.6 to 78.3 and IDF1 from 79.3 to 80.2, when $\sigma$ is 20. Tracklet interpolation is an effective post-processing method to obtain the boxes of those fully-occluded objects. We use tracklet interpolation in the test sets of MOT17 \cite{milan2016mot16}, MOT20 \cite{dendorfer2020mot20} and HiEve \cite{lin2020human} under the private detection protocol.

\section{Public detection results on MOTChallenge}
We evaluate ByteTrack on the test set of MOT17 \cite{milan2016mot16} and MOT20 \cite{dendorfer2020mot20} under the public detection protocol. Following the public detection filtering strategy in Tracktor \cite{bergmann2019tracking} and CenterTrack \cite{zhou2020tracking}, we only initialize a new trajectory when its IoU with a public detection box is larger than 0.8. We do not use tracklet interpolation under the public detection protocol. As is shown in Table~\ref{table_mot17pub}, ByteTrack outperforms other methods by a large margin on MOT17. For example, it outperforms SiamMOT by 1.5 points on MOTA and 6.7 points on IDF1. Table~\ref{table_mot20pub} shows the results on MOT20. ByteTrack also outperforms existing results by a large margin. For example, it outperforms TMOH \cite{stadler2021improving} by 6.9 points on MOTA, 9.0 points on IDF1, 7.5 points on HOTA and reduce the identity switches by three quarters. The results under public detection protocol further indicate the effectiveness of our association method BYTE.

\begin{table}[t]
\begin{center}
\scalebox{0.85}{\input{tables/mot17_pub}}
\end{center}
\vspace{-6mm}
\caption{Comparison of the state-of-the-art methods under the “public detector” protocol on MOT17 test set. The best results are shown in \textbf{bold}. }
\label{table_mot17pub}
\vspace{-4mm}
\end{table}

\begin{table}[t]
\begin{center}
\scalebox{0.85}{\input{tables/mot20_pub}}
\end{center}
\vspace{-6mm}
\caption{Comparison of the state-of-the-art methods under the “public detector” protocol on MOT20 test set. The best results are shown in \textbf{bold}. }
\label{table_mot20pub}
\vspace{-4mm}
\end{table}

\section{Visualization results. }
\begin{figure*}[!h]
	\centering
	\includegraphics[width=0.65\linewidth]{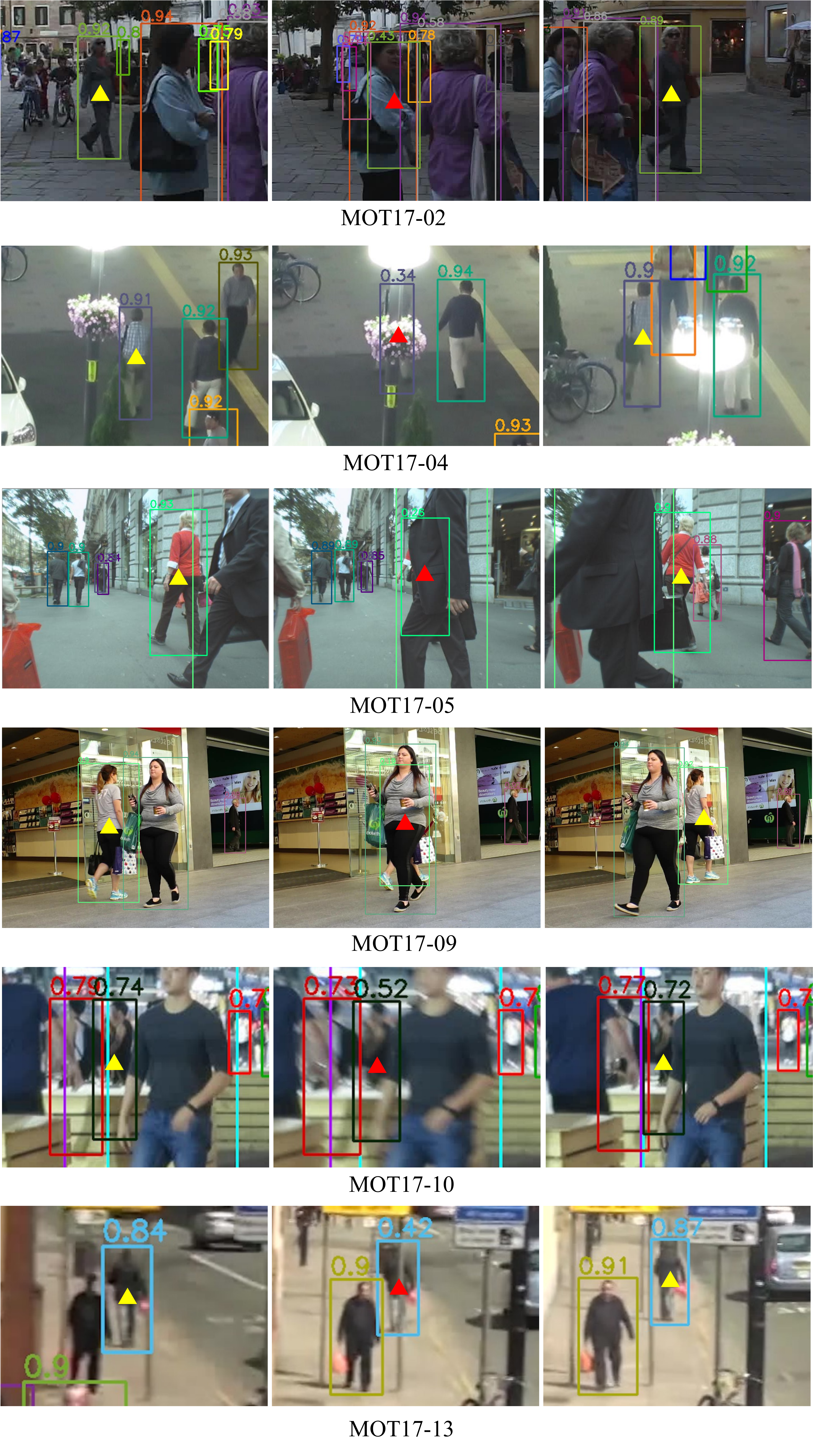}
	\caption{Visualization results of ByteTrack. We select 6 sequences from the validation set of MOT17 and show the effectiveness of ByteTrack to handle difficult cases such as occlusion and motion blur. The yellow triangle represents the high score box and the red triangle represents the low score box. The same box color represents the same identity. }
	\label{fig:vis}
	\vspace{-2mm}
\end{figure*}

We show some visualization results of difficult cases which ByteTrack is able to handle in Figure~\ref{fig:vis}. The difficult cases include occlusion (\ie MOT17-02, MOT17-04, MOT17-05, MOT17-09, MOT17-13), motion blur (\ie MOT17-10, MOT17-13) and small objects (\ie MOT17-13). The pedestrian in the middle frame with red triangle has low detection score, which is obtained by our association method BYTE. The low score boxes not only decrease the number of missing detection, but also play an important role for long-range association. As we can see from all these difficult cases, ByteTrack does not bring any identity switch and preserve the identity effectively.

\clearpage
{\small
\bibliographystyle{ieee}
\bibliography{egbib}
}

\end{document}

%% file: tables/sim.tex
\setlength{\tabcolsep}{12.0pt}

\begin{tabular}{ l l c c c c c c}

\toprule
 & & & MOT17 & & & BDD100K & \\
Similarity\#1 & Similarity\#2 & MOTA$\uparrow$ & IDF1$\uparrow$ & IDs$\downarrow$ & mMOTA$\uparrow$ & mIDF1$\uparrow$ & IDs$\downarrow$\\
\cmidrule(lr){1-2} \cmidrule(lr){3-5} \cmidrule(lr){6-8}
IoU & Re-ID & 75.8 & 77.5 & 231 & 39.2 & 48.3 & 29172\\
IoU & IoU & \textbf{76.6} & 79.3 & \textbf{159} & 39.4 & 48.9 & 27902\\
Re-ID & Re-ID & 75.2 & 78.7 & 276 & 45.0 & 53.4 & 10425\\
Re-ID & IoU & 76.3 & \textbf{80.5} & 216 & \textbf{45.5} & \textbf{54.8} & \textbf{9140}\\
\bottomrule
\end{tabular}

%% file: tables/ass.tex
\setlength{\tabcolsep}{10pt}
\begin{tabular}{ l  c  c c c  c c c  c}
\toprule
& & & MOT17 & &  & BDD100K & & \\
Method & w/ Re-ID & MOTA$\uparrow$ & IDF1$\uparrow$ & IDs$\downarrow$ & mMOTA$\uparrow$ & mIDF1$\uparrow$ & IDs$\downarrow$ & FPS \\
\cmidrule(lr){1-2} \cmidrule(lr){3-5} \cmidrule(lr){6-8} \cmidrule(lr){9-9}
SORT &  & 74.6 & 76.9 & 291 & 30.9 & 41.3 & 10067 & \textbf{30.1}\\
DeepSORT & $\checkmark$ & 75.4 & 77.2 & 239 & 24.5 & 38.2 & 10720 & 13.5\\
MOTDT & $\checkmark$ & 75.8 & 77.6 & 273 & 26.7 & 39.8 & 14520 & 11.1\\
\textbf{BYTE (ours)} &  & \textbf{76.6} & 79.3 & \textbf{159} & 39.4 & 48.9 & 27902 & 29.6 \\
\textbf{BYTE (ours)} & $\checkmark$  & 76.3 & \textbf{80.5} & 216 & \textbf{45.5} & \textbf{54.8} & \textbf{9140} & 11.8 \\
\bottomrule
\end{tabular}

%% file: tables/app.tex
\setlength{\tabcolsep}{2.0pt}

\begin{tabular}{ l | c c | l l c c l }

\toprule
Method & Similarity & w/ BYTE & MOTA$\uparrow$ & IDF1$\uparrow$ & IDs$\downarrow$\\
\midrule
JDE \cite{wang2020towards} & Motion(K) + Re-ID &  & 60.0 & 63.6 & 473\\
 & Motion(K) + Re-ID & $\checkmark$ & 60.3 (+0.3) & 64.1 (+0.5) & 418\\
 & Motion(K) & $\checkmark$ & 60.6 (+0.6) & 66.0 \textcolor{codegreen}{(+\textbf{2.4})} & 360\\

\midrule
CSTrack \cite{liang2020rethinking} & Motion(K) + Re-ID &  & 68.0 & 72.3 & 325\\
 & Motion(K) + Re-ID & $\checkmark$ & 69.2 \textcolor{codegreen}{(+\textbf{1.2})} & 73.9 \textcolor{codegreen}{(+\textbf{1.6})} & 285\\
 & Motion(K) & $\checkmark$ & 69.3 \textcolor{codegreen}{(+\textbf{1.3})} & 71.7 (-0.6) & 279\\

\midrule
FairMOT \cite{zhang2020fairmot} & Motion(K) + Re-ID &  & 69.1 & 72.8 & 299\\
 & Motion(K) + Re-ID & $\checkmark$ & 70.4 \textcolor{codegreen}{(+\textbf{1.3})} & 74.2 \textcolor{codegreen}{(+\textbf{1.4})} & 232\\
 & Motion(K) & $\checkmark$ & 70.3 \textcolor{codegreen}{(+\textbf{1.2})} & 73.2 (+0.4) & 236\\
 
\midrule
TraDes \cite{wu2021track} & Motion + Re-ID &  & 68.2 & 71.7 & 285\\
 & Motion + Re-ID & $\checkmark$ & 68.6 (+0.4) & 71.1 (-0.6) & 259\\
 & Motion(K) & $\checkmark$ & 67.9 (-0.3) & 72.0 (+0.3) & 178\\

\midrule
QuasiDense \cite{pang2021quasi} & Re-ID & & 67.3 & 67.8 & 377\\
& Motion(K) + Re-ID & $\checkmark$ & 67.7 (+0.4) & 72.0 \textcolor{codegreen}{(+\textbf{4.2})} & 281\\
& Motion(K) & $\checkmark$ & 67.9  (+0.6) & 70.9  \textcolor{codegreen}{(+\textbf{3.1})} & 258\\
 
\midrule
CenterTrack \cite{zhou2020tracking} & Motion &  & 66.1 & 64.2 & 528\\
 & Motion & $\checkmark$ & 66.3 (+0.2) & 64.8 (+0.6) & 334\\
 & Motion(K) & $\checkmark$ & 67.4 \textcolor{codegreen}{(+\textbf{1.3})} & 74.0 \textcolor{codegreen}{(+\textbf{9.8})} & 144\\
 
\midrule
CTracker \cite{peng2020chained} & Chain &  & 63.1 & 60.9 & 755\\
 & Motion(K) & $\checkmark$ & 65.0 \textcolor{codegreen}{(+\textbf{1.9})} & 66.7 \textcolor{codegreen}{(+\textbf{5.8})} & 346\\

\midrule
TransTrack \cite{sun2020transtrack} & Attention &  & 67.1 & 68.3 & 254\\
 & Attention & $\checkmark$ & 68.6 \textcolor{codegreen}{(+\textbf{1.5})} & 69.0 (+0.7) & 232\\
 & Motion(K) & $\checkmark$ & 68.3 \textcolor{codegreen}{(+\textbf{1.2})}& 72.4  \textcolor{codegreen}{(+\textbf{4.1})} & 181\\
 
\midrule
MOTR \cite{zeng2021motr} & Attention &  & 64.7 & 67.2 & 346\\
 & Attention & $\checkmark$ & 64.3 (-0.4) & 69.3 \textcolor{codegreen}{(+\textbf{2.1})} & 263\\
 & Motion(K) & $\checkmark$ & 65.7 \textcolor{codegreen}{(+\textbf{1.0})} & 68.4 \textcolor{codegreen}{(+\textbf{1.2})} & 260\\
\bottomrule
\end{tabular}

%% file: tables/mot17.tex
\setlength{\tabcolsep}{2.0pt}

\begin{tabular}{ l | c c c c c c c}

\toprule
Tracker & MOTA$\uparrow$ & IDF1$\uparrow$ & HOTA$\uparrow$ & FP$\downarrow$ & FN$\downarrow$ & IDs$\downarrow$ & FPS$\uparrow$\\
\midrule
DAN \cite{sun2019deep} & 52.4 & 49.5 & 39.3 & 25423 & 234592 & 8431 & \textless 3.9\\ 
Tube\_TK \cite{pang2020tubetk} & 63.0 & 58.6 & 48.0 & 27060 & 177483 & 4137 & 3.0\\
MOTR \cite{zeng2021motr} & 65.1 & 66.4 & - & 45486 & 149307 & 2049 & -\\
CTracker \cite{peng2020chained} & 66.6 & 57.4 & 49.0 & 22284 & 160491 & 5529 & 6.8\\
CenterTrack \cite{zhou2020tracking} & 67.8 & 64.7 & 52.2 & \textbf{18498} & 160332 & 3039 & 17.5\\
QuasiDense \cite{pang2021quasi} & 68.7 & 66.3 & 53.9 & 26589 & 146643 & 3378 & 20.3\\
TraDes \cite{wu2021track} & 69.1 & 63.9 & 52.7 & 20892 & 150060 & 3555 & 17.5\\
MAT \cite{han2020mat} & 69.5 & 63.1 & 53.8 & 30660 & 138741 & 2844 & 9.0\\
SOTMOT \cite{zheng2021improving} & 71.0 & 71.9 & - & 39537 & 118983 & 5184 & 16.0\\
TransCenter \cite{xu2021transcenter} & 73.2 & 62.2 & 54.5 & 23112 & 123738 & 4614 & 1.0\\
GSDT \cite{wang2020joint} & 73.2 & 66.5 & 55.2 & 26397 & 120666 & 3891 & 4.9\\
Semi-TCL \cite{li2021semi} & 73.3 & 73.2 & 59.8 & 22944 & 124980 & 2790 & -\\
FairMOT \cite{zhang2020fairmot} & 73.7 & 72.3 & 59.3 & 27507 & 117477 & 3303 & 25.9\\
RelationTrack \cite{yu2021relationtrack} & 73.8 & 74.7 & 61.0 & 27999 & 118623 & \textbf{1374} & 8.5\\
PermaTrackPr \cite{tokmakov2021learning} & 73.8 & 68.9 & 55.5 & 28998 & 115104 & 3699 & 11.9\\
CSTrack \cite{liang2020rethinking} & 74.9 & 72.6 & 59.3 & 23847 & 114303 & 3567 & 15.8\\
TransTrack \cite{sun2020transtrack} & 75.2 & 63.5 & 54.1 & 50157 & 86442 & 3603 & 10.0\\
FUFET \cite{shan2020tracklets} & 76.2 & 68.0 & 57.9 & 32796 & 98475 & 3237 & 6.8\\
SiamMOT \cite{liang2021one} & 76.3 & 72.3 & - & - & - & - & 12.8\\
CorrTracker \cite{wang2021multiple} & 76.5 & 73.6 & 60.7 & 29808 & 99510 & 3369 & 15.6\\
TransMOT \cite{chu2021transmot} & 76.7 & 75.1 & 61.7 & 36231 & 93150 & 2346 & 9.6\\
ReMOT \cite{yang2021remot} & 77.0 & 72.0 & 59.7 & 33204 & 93612 & 2853 & 1.8\\
\textbf{ByteTrack (ours)} & \textbf{80.3} & \textbf{77.3} & \textbf{63.1} & 25491 & \textbf{83721} & 2196 & \textbf{29.6}\\
\bottomrule
\end{tabular}

%% file: tables/mot20.tex
\setlength{\tabcolsep}{1.5pt}

\begin{tabular}{ l | c c c c c c c}

\toprule
Tracker & MOTA$\uparrow$ & IDF1$\uparrow$ & HOTA$\uparrow$ & FP$\downarrow$ & FN$\downarrow$ & IDs$\downarrow$ & FPS$\uparrow$\\
\midrule
MLT \cite{zhang2020multiplex} & 48.9 & 54.6 & 43.2 & 45660 & 216803 & 2187 & 3.7\\
FairMOT \cite{zhang2020fairmot} & 61.8 & 67.3 & 54.6 & 103440 & 88901 & 5243 & 13.2\\
TransCenter \cite{xu2021transcenter} & 61.9 & 50.4 & - & 45895 & 146347 & 4653 & 1.0\\
TransTrack \cite{sun2020transtrack} & 65.0 & 59.4 & 48.5 & 27197 & 150197 & 3608 & 7.2\\
CorrTracker \cite{wang2021multiple} & 65.2 & 69.1 & - & 79429 & 95855 & 5183 & 8.5\\
Semi-TCL \cite{li2021semi} & 65.2 & 70.1 & 55.3 & 61209 & 114709 & 4139 & -\\
CSTrack \cite{liang2020rethinking} & 66.6 & 68.6 & 54.0 & \textbf{25404} & 144358 & 3196 & 4.5\\
GSDT \cite{wang2020joint} & 67.1 & 67.5 & 53.6 & 31913 & 135409 & 3131 & 0.9\\
SiamMOT \cite{liang2021one} & 67.1 & 69.1 & - & - & - & - & 4.3\\
RelationTrack \cite{yu2021relationtrack} & 67.2 & 70.5 & 56.5 & 61134 & 104597 & 4243 & 2.7\\
SOTMOT \cite{zheng2021improving} & 68.6 & 71.4 & - & 57064 & 101154 & 4209 & 8.5\\
\textbf{ByteTrack (ours)} & \textbf{77.8} & \textbf{75.2} & \textbf{61.3} & 26249 & \textbf{87594} & \textbf{1223} & \textbf{17.5}\\
\bottomrule
\end{tabular}

%% file: tables/hie.tex
\setlength{\tabcolsep}{2.5pt}

\begin{tabular}{ l | c c c c c c c}

\toprule
Tracker & MOTA$\uparrow$ & IDF1$\uparrow$ & MT$\uparrow$ & ML$\downarrow$ & FP$\downarrow$ & FN$\downarrow$ & IDs$\downarrow$\\
\midrule
DeepSORT \cite{wojke2017simple} & 27.1 & 28.6 & 8.5\% & 41.5\% & 5894 & 42668 & 2220\\
MOTDT \cite{chen2018real} & 26.1 & 32.9 & 8.7\% & 54.6\% & 6318 & 43577 & 1599\\
IOUtracker \cite{bochinski2017high} & 38.6 & 38.6 & 28.3\% & 27.6\% & 9640 & 28993 & 4153\\
JDE \cite{wang2020towards} & 33.1 & 36.0 & 15.1\% & 24.1\% & 9526 & 33327 & 3747\\
FairMOT \cite{zhang2020fairmot} & 35.0 & 46.7 & 16.3\% & 44.2\% & 6523 & 37750 & \textbf{995}\\
CenterTrack \cite{zhou2020tracking} & 40.9 & 45.1 & 10.8\% & 32.2\% & 3208 & 36414 & 1568\\
\textbf{ByteTrack (Ours)} & \textbf{61.7} & \textbf{63.1} & \textbf{38.3\%} & \textbf{21.6\%} & \textbf{2822} & \textbf{22852} & 1031\\
\bottomrule
\end{tabular}

%% file: tables/bdd100k.tex
\setlength{\tabcolsep}{4pt}

\begin{tabular}{l cccccccccc}
\toprule
Tracker & split & mMOTA$\uparrow$ & mIDF1$\uparrow$ & MOTA$\uparrow$ & IDF1$\uparrow$ & FN$\downarrow$ & FP$\downarrow$ & IDs$\downarrow$ & MT$\uparrow$ & ML$\downarrow$ \\
\midrule
Yu \etal \cite{yu2020bdd100k} & val & 25.9 & 44.5 & 56.9 & 66.8 & 122406 & 52372 & 8315 & 8396 & 3795\\
QDTrack \cite{pang2021quasi} & val & 36.6 & 50.8 & 63.5 & \textbf{71.5} & 108614 & 46621 & \textbf{6262} & 9481 & 3034\\
\textbf{ByteTrack(Ours)} & val & \textbf{45.5} & \textbf{54.8} & \textbf{69.1} & 70.4 & \textbf{92805} & \textbf{34998} & 9140 & \textbf{9626} & \textbf{3005}\\
\midrule
Yu \etal \cite{yu2020bdd100k} & test & 26.3 & 44.7 & 58.3 & 68.2 & 213220 & 100230 & 14674  & 16299 & 6017\\

DeepBlueAI & test & 31.6 & 38.7 & 56.9& 56.0& 292063 & \textbf{35401} & 25186  & 10296 & 12266\\

madamada & test & 33.6 & 43.0 & 59.8& 55.7& 209339 & 76612  & 42901  & 16774 & \textbf{5004}\\

QDTrack\cite{pang2021quasi} & test & 35.5 & 52.3 & 64.3 & \textbf{72.3}  & 201041 & 80054  & \textbf{10790} & 17353 & 5167\\

\textbf{ByteTrack(Ours)} & test & \textbf{40.1} & \textbf{55.8} & \textbf{69.6} & 71.3 & \textbf{169073} & 63869 & 15466 & \textbf{18057} & 5107\\
\bottomrule
\end{tabular}

%% file: tables/light.tex
\setlength{\tabcolsep}{4pt}

\begin{tabular}{ l c c | l c c c}

\toprule
Backbone & Params & GFLOPs & Tracker & MOTA$\uparrow$ & IDF1$\uparrow$ & IDs$\downarrow$ \\
\midrule
YOLOX-M & 25.3 M & 118.7 & DeepSORT & 74.5 & 76.2 & 197\\
YOLOX-M & 25.3 M & 118.7 & BYTE & 75.3 & 77.5 & 200\\
YOLOX-S & 8.9 M & 43.0 & DeepSORT & 69.6 & 71.5 & 205\\
YOLOX-S & 8.9 M & 43.0 & BYTE & 71.1 & 73.6 & 224\\
YOLOX-Tiny & 5.0 M & 24.5 & DeepSORT & 68.6 & 72.0 & 224\\
YOLOX-Tiny & 5.0 M & 24.5 & BYTE & 70.5 & 72.1 & 222\\
YOLOX-Nano & 0.9 M & 4.0 & DeepSORT & 61.4 & 66.8 & 212\\
YOLOX-Nano & 0.9 M & 4.0 & BYTE & 64.4 & 68.4 & 161\\
\bottomrule
\end{tabular}

%% file: tables/input_size.tex
\setlength{\tabcolsep}{5pt}

\begin{tabular}{ l | c c c c}

\toprule
Input size & MOTA$\uparrow$ & IDF1$\uparrow$ & IDs$\downarrow$ & Time (ms) \\
\midrule
$512 \times 928$ & 75.0 & 77.6 & 200 & \textbf{17.9+4.0}\\
$608 \times 1088$ & 75.6 & 76.4 & 212 & 21.8+4.0\\
$736 \times 1280$ & 76.2 & 77.4 & 188 & 26.2+4.2\\
$800 \times 1440$ & \textbf{76.6} & \textbf{79.3} & \textbf{159} & 29.6+4.2 \\
\bottomrule
\end{tabular}

%% file: tables/training_data.tex
\setlength{\tabcolsep}{4pt}

\begin{tabular}{ l | c | c c c}

\toprule
Training data & Images & MOTA$\uparrow$ & IDF1$\uparrow$ & IDs$\downarrow$\\
\midrule
MOT17 & 2.7K & 75.8 & 76.5 & 205\\
MOT17 + CH & 22.0K & 76.6 & 79.3 & \textbf{159}\\
MOT17 + CH + CE & 26.6K & \textbf{76.7} & \textbf{79.7} & 183 \\
\bottomrule
\end{tabular}

%% file: tables/inter.tex
\setlength{\tabcolsep}{6pt}

\begin{tabular}{ c | c c c c c c}

\toprule
Interval & MOTA$\uparrow$ & IDF1$\uparrow$ & FP$\downarrow$ & FN$\downarrow$ & IDs$\downarrow$\\
\midrule
No & 76.6 & 79.3 & \textbf{3358} & 9081 & 159\\
10 & 77.4 & 79.7 & 3638 & 8403 & 150\\
20 & \textbf{78.3} & \textbf{80.2} & 3941 & 7606 & \textbf{146}\\
30 & 78.3 & 80.2 & 4237 & \textbf{7337} & 147\\
\bottomrule
\end{tabular}

%% file: tables/mot17_pub.tex
\setlength{\tabcolsep}{2.0pt}

\begin{tabular}{ l | c c c c c c}

\toprule
Tracker & MOTA$\uparrow$ & IDF1$\uparrow$ & HOTA$\uparrow$ & FP$\downarrow$ & FN$\downarrow$ & IDs$\downarrow$\\
\midrule
STRN \cite{xu2019spatial} & 50.9 & 56.0 & 42.6 & 25295 & 249365 & 2397\\
FAMNet \cite{chu2019famnet} & 52.0 & 48.7 & - & 14138 & 253616 & 3072\\
Tracktor++v2 \cite{bergmann2019tracking} & 56.3 & 55.1 & 44.8 & \textbf{8866} & 235449 & 1987\\
MPNTrack \cite{braso2020learning} & 58.8 & 61.7 & 49.0 & 17413 & 213594 & 1185\\
LPC\_MOT \cite{dai2021learning} & 59.0 & 66.8 & 51.5 & 23102 & 206948 & \textbf{1122}\\
Lif\_T \cite{hornakova2020lifted} & 60.5 & 65.6 & 51.1 & 14966 & 206619 & 1189\\
CenterTrack \cite{zhou2020tracking} & 61.5 & 59.6 & 48.2 & 14076 & 200672 & 2583\\
TMOH \cite{stadler2021improving} & 62.1 & 62.8 & 50.4 & 10951 & 201195 & 1897\\
ArTIST\_C \cite{saleh2021probabilistic} & 62.3 & 59.7 & 48.9 & 19611 & 191207 & 2062\\
QDTrack \cite{pang2021quasi} & 64.6 & 65.1 & - & 14103 & 182998 & 2652\\
SiamMOT \cite{shuai2021siammot} & 65.9 & 63.3 & - & 18098 & 170955 & 3040\\
\textbf{ByteTrack (ours)} & \textbf{67.4} & \textbf{70.0} & \textbf{56.1} & 9939 & \textbf{172636} & 1331\\
\bottomrule
\end{tabular}

%% file: tables/mot20_pub.tex
\setlength{\tabcolsep}{2.0pt}

\begin{tabular}{ l | c c c c c c}

\toprule
Tracker & MOTA$\uparrow$ & IDF1$\uparrow$ & HOTA$\uparrow$ & FP$\downarrow$ & FN$\downarrow$ & IDs$\downarrow$\\
\midrule
SORT \cite{bewley2016simple} & 42.7 & 45.1 & 36.1 & 27521 & 264694 & 4470\\
Tracktor++v2 \cite{bergmann2019tracking} & 52.6 & 52.7 & 42.1 & \textbf{6930} & 236680 & 1648\\
ArTIST\_C \cite{saleh2021probabilistic} & 53.6 & 51.0 & 41.6 & 7765 & 230576 & 1531\\
LPC\_MOT \cite{dai2021learning} & 56.3 & 62.5 & 49.0 & 11726 & 213056 & 1562\\
MPNTrack \cite{braso2020learning} & 57.6 & 59.1 & 46.8 & 16953 & 201384 & 1210\\
TMOH \cite{stadler2021improving} & 60.1 & 61.2 & 48.9 & 38043 & 165899 & 2342\\
\textbf{ByteTrack (ours)} & \textbf{67.0} & \textbf{70.2} & \textbf{56.4} & 9685 & \textbf{160303} & \textbf{680}\\
\bottomrule
\end{tabular}